 \documentclass[times, review, 10pt]{elsarticle}

\usepackage{wrapfig}
\usepackage{flushend}
\usepackage{amssymb}
\usepackage{ulem}
\usepackage{amsmath}
\usepackage{multirow}
\journal{Pattern Recognition}

\begin{document}

\begin{frontmatter}



\title{LGD: Leveraging Generative Descriptions for Zero-Shot Referring Image Segmentation}

\author[1]{Jiachen Li}
\author[1,2]{Qing Xie\corref{cor1}}
\ead{felixxq@whut.edu.cn}
\author[3]{Renshu Gu\corref{cor1}}
\ead{renshugu@hdu.edu.cn}
\author[1]{Jinyu Xu}
\author[1,2]{Yongjian Liu}
\author[3]{Xiaohan Yu}
\cortext[cor1]{Corresponding author}

\affiliation[1]{organization={School of Computer Science and Artificial Intelligence, Wuhan University of Technology},
            addressline={},
            city={Wuhan},
            postcode={430070}, 
            state={Hubei},
            country={China}}
\affiliation[2]{organization={Engineering Research Center of Intelligent Service Technology for Digital Publishing , Ministry of Education},
    city={Wuhan},
    postcode={430070}, 
    state={Hubei},
    country={China}}
\affiliation[3]{organization={School of Computer Science, Hangzhou Dianzi University},
    city={Hangzhou},
    postcode={310018}, 
    state={Zhejiang},
    country={China}}
\affiliation[4]{organization={School of Computing, Macquarie University},
    city={Sydney},
    postcode={2109}, 
    state={NSW},
    country={Australia}}

\begin{abstract}
Zero-shot referring image segmentation aims to locate and segment the target region based on a referring expression, with the primary challenge of aligning and matching semantics across visual and textual modalities without training.
Previous works address this challenge by utilizing Vision-Language Models and mask proposal networks for region-text matching.
However, this paradigm may lead to incorrect target localization due to the inherent ambiguity and diversity of free-form referring expressions.
To alleviate this issue, we present LGD (Leveraging Generative Descriptions), a framework that utilizes the advanced language generation capabilities of Multi-Modal Large Language Models to enhance region-text matching performance in Vision-Language Models.
Specifically, we first design two kinds of prompts, the attribute prompt and the surrounding prompt, to guide the Multi-Modal Large Language Models in generating descriptions related to the crucial attributes of the referent object and the details of surrounding objects, referred to as attribute description and surrounding description, respectively.
Secondly, three visual-text matching scores are introduced to evaluate the similarity between instance-level visual features and textual features, which determines the mask most associated with the referring expression.
The proposed method achieves new state-of-the-art performance on three public datasets RefCOCO, RefCOCO+ and RefCOCOg, with maximum improvements of 9.97\% in oIoU and 11.29\% in mIoU compared to previous methods.
\end{abstract}


\begin{highlights}
\item Introduce LGD, a novel framework that leverages generative descriptions from Multi-Modal Large Language Models to enhance visual-text alignment in zero-shot referring image segmentation.
\item Design two types of prompts—attribute and surrounding prompts—to guide the generation of fine-grained and context-aware textual descriptions, improving region-text matching accuracy.
\item Set new state-of-the-art results on RefCOCO, RefCOCO+, and RefCOCOg datasets, with up to 9.97\% oIoU and 11.29\% mIoU improvements over previous approaches.

\end{highlights}


\end{frontmatter}



\section{Introduction}
\label{sec1}

Referring Image Segmentation (RIS) aims to identify and segment a specific object within an image through a free-form referring expression, presenting a foundational and long-standing challenge in multi-modal understanding.
Compared to traditional visual segmentation, such as semantic segmentation~\cite{ss1,ss2,ss3,ss4}, instance segmentation~\cite{is1,is2,is3}, and panoptic segmentation~\cite{ps1}, RIS locates object without relying on a fixed set of predefined categories.
Additionally, unlike open-vocabulary segmentation~\cite{ovs1}, which identifies objects based on a novel set of categories, RIS requires accurate understanding of referring expressions to locate objects, due to the inherent ambiguity and diversity in such text descriptions.
RIS demonstrates flexibility and adaptability, with broad potential for applications such as image editing~\cite{app11}, vision-language navigation~\cite{app21}, and visual question answering~\cite{app31}.
Although current RIS methods~\cite{cris-ris,vlt-ris,lavt-ris,groupformer-ris} achieve strong performance, they depend on pixel-level masks that require extensive manual annotation to train effectively.
Weakly supervised RIS methods~\cite{kim-wris,lee-wris,ppt-wris} provides an alternative by utilizing text, points, or bounding boxes instead of dense mask annotations, however, it still requires labor-intensive annotations and cumbersome training pipelines.

The intensive labor costs associated with manual annotations limit the development of RIS, and the zero-shot learning is utilized to alleviate the problem. 
Zero-shot RIS leverages Vision-Language Models (\textit{e.g.}, CLIP~\cite{CLIP}) to effectively match referring expressions with instance-level mask from off-the-shelf mask proposal networks (\textit{e.g.}, SAM~\cite{SAM} and FreeSOLO~\cite{Freesolo}), avoiding the requirement for manual pixel-level annotations and traditional training processes.
Global-Local CLIP~\cite{GLCLIP2023} is the first proposed zero-shot RIS method. This framework employs FreeSOLO to obtain fine-grained instance-level grounding, then utilizes CLIP to extract both global and local contextual information from instance-level masks and referring expressions, and finally selects the target mask based on the similarity of the global-local mixed features.
TAS~\cite{TAS2023} and BMS~\cite{BMS} design various visual-text matching scores to comprehensively evaluate the fine-grained cross-modal correlation between instance masks and referring expressions.

\begin{figure}[!t]
\centering
\includegraphics[width=0.75\linewidth]{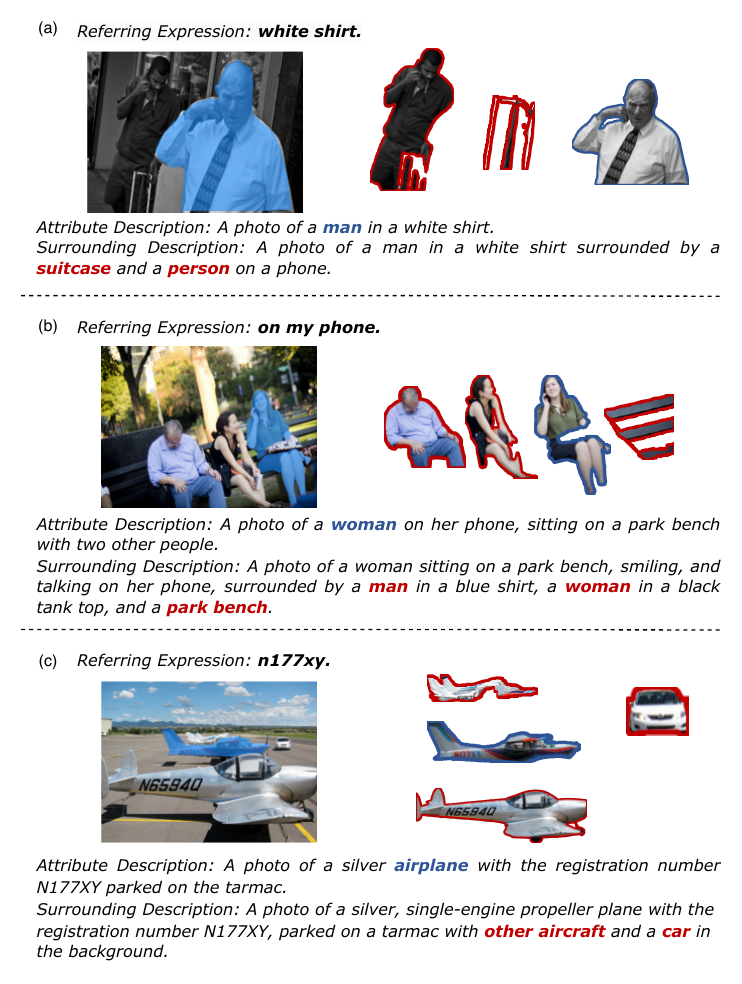}
\caption{
{\bf An illustration demonstrating the inherent ambiguity and diversity of free-form referring expressions.}
(a) The referring expression describes the clothing of the referent object without explicitly mentioning the man wearing a white shirt.
(b) The referring expression highlights the action of the referent object rather than explicitly describing the woman making the call.
(c) The referring expression conveys the registration number of the referent object rather than explicitly mentioning the aircraft registered as n177xy.
Referring expressions are unrestricted in describing object and lack detailed descriptions of referent object, making it challenging for the model to accurately locate the referent object.
We introduced attribute description and surrounding description, which assist in identifying the referent object and its crucial attributes while distinguishing it from surrounding objects, providing the model with fine-grained information.}
\label{fig_intro}
\end{figure}

The aforementioned methods leverage the powerful image-text matching capabilities of Vision-Language Models to calculate region-text similarity. However, their performance remains constrained by the inherent ambiguity and diversity of free-form referring expressions, which make it difficult to align textual entities with visual regions.
Specifically, referring expressions are unrestricted in describing objects and inevitably lack detailed descriptions of referent object, which limits the capacity of the model to achieve fine-grained understanding of the referent object.
For example, as shown in Fig.~\ref{fig_intro}, (a), (b) and (c) describe the clothing, action, and registration number of the referent object, respectively, without including entity category, which increases the difficulty of accurately locating the referent object.
Furthermore, referring expressions tend to be inadequate in describing all object relations in the input image, as shown in Fig.~\ref{fig_intro}(a), (b), (c), which restricts a holistic understanding of multi-modal relational information and limits the ability to distinguish the referent object from others.

To address these challenges, we propose LGD, a framework designed to leverage the powerful language generation capabilities of Multi-Modal Large Language Models (MLLMs) to enhance the region-text matching performance of Vision-Language Models.
Our approach introduces two key prompts: the attribute prompt and the surrounding prompt, which together guide the MLLMs in extracting fine-grained contextual information.
Specifically, the attribute prompt enables the extraction of crucial attribute description of the referent object, providing a more comprehensive and detailed characterization of the referent object, addressing the limitations of the referring expression in comprehensive description.
On the other hand, the surrounding prompt guides the MLLMs to generate descriptions of objects around the referent object, which aids in distinguishing the referent object from others in the scene.
Unlike referring expressions, which are inherently ambiguous and diverse, attribute and surrounding descriptions offer an alternative perspective by providing more comprehensive and discriminative information about the referent object.
Similar to TAS~\cite{TAS2023} and BMS~\cite{BMS}, we further utilize the surrounding object information as negative noun phrases to refine the distinction between the referent and non-referent objects.
Inspired by existing zero-shot RIS methods~\cite{GLCLIP2023,TAS2023,BMS}, we compute three visual-text matching scores using CLIP: (1) the similarity between the instance-level visual feature and the crucial attribute description, denoted as $S_{att}$, (2) the similarity between the instance-level visual feature and the surrounding object noun phrases, denoted as $S_{sur}$, and (3) the similarity between the instance-level visual feature and the referring expression, denoted as $S_{van}$.
Finally, a linear combination of these three scores determines the mask most closely aligned with the referring expression, ensuring accurate identification of the referent object.

Our work is summarized as follows:
\begin{itemize}
\item{Our method incorporates two prompts, the attribute prompt and the surrounding prompt, to guide MLLMs in generating detailed descriptions of the referent objects, effectively addressing the limitations of free-form referring expressions by providing more comprehensive and precise object representations.}
\item{We propose three visual-text matching scores, $S_{att}$, $S_{sur}$, and $S_{van}$, to improve the alignment between instance-level visual features and referring expressions.}
\item{Our method surpasses existing zero-shot RIS methods, achieving state-of-the-art results on the RefCOCO, RefCOCO+, and RefCOCOg.}
\end{itemize}

\section{Related Work}
\subsection{Zero-Shot Referring Image Segmentation}
Fully supervised RIS~\cite{cris-ris,vlt-ris,lavt-ris,groupformer-ris} and weakly supervised RIS \cite{kim-wris,lee-wris,ppt-wris} methods face significant limitations in practical application scenarios due to their reliance on large amounts of labeled data and the high training costs. To address these challenges, zero-shot RIS methods leverage the zero-shot matching capabilities of Vision-Language Models to align pixel-level regions with referring expressions.
Yu \textit{et al.}~\cite{GLCLIP2023} introduced the first zero-shot RIS method termed Global-Local CLIP. This method utilizes a visual encoder and a text encoder to extract global and local contextual information from mask-guided images and referring expressions. By leveraging the cross-modal knowledge of the pre-trained CLIP, it achieves accurate matching between instance-level masks and referring expressions.
Suo~\textit{et al.} proposed TAS~\cite{TAS2023}, which utilizes BLIP2~\cite{blip2} to generate captions for input images and introduces a text-augmented visual-text matching score to enhance the region-text aligning ability of CLIP.
Li~\textit{et al.}~\cite{BMS} proposed the Bidirectional Mask Selection (BMS) framework, which enhances cross-modal fine-grained correlation understanding by leveraging the complementarity between positive and negative masks, achieving accurate fine-grained matching between instance-level masks and referring expressions.
Sun \textit{et al.} proposed CaR~\cite{clipasrnn}, which iteratively refines the alignment between referring expressions and predicted masks, ultimately generating accurate segmentation results.
Furthermore, several zero-shot referring expression comprehension~\cite{zrec1,zrec2} methods have been developed to adapt pre-trained Vision-Language Models. 
In contrast to the aforementioned methods, our approach leverages MLLMs to comprehensively understand the crucial attributes of the referent object and surrounding objects, enhancing the alignment of Vision-Language Models with instance-level masks and referring expressions.

\subsection{MLLMs for Language-Driven Image Segmentation}
Language-driven image segmentation has gained significant attention due to its potential to enable fine-grained object localization through natural language instructions.
Early methods \cite{EFN, lavt-ris} focused on multi-modal feature fusion and semantic alignment to achieve competitive performance.
Recent advancements in MLLMs \cite{llava,gpt4} have introduced a paradigm shift by incorporating extensive pretraining on multi-modal data and leveraging the generative capabilities of language models.
Lisa \cite{lisa} and Gsva \cite{gsva} utilize the \textit{[SEG]} tokens generated by MLLMs as a prompt to guide SAM in producing the corresponding mask.
Similarly, PixelLM \cite{PixelLM} combines MLLMs with a codebook to design the \textit{[SEG]} tokens, demonstrating its effectiveness in a range of pixel-level reasoning and understanding tasks.
SAM4MLLM \cite{chen2025sam4MLLM} employs MLLMs to generate proposal points for referent object through both direct and indirect methods.
OMG-LLaVA \cite{zhang2024omg} leverages visual tokens that integrate multi-modal information to guide the LLM in understanding text instructions and generating \textit{[SEG]} tokens for multi-level segmentation tasks.
In contrast to these methods, our LGD adopts an intuitive and simple way to directly transfer knowledge from MLLMs using the designed attribute prompt and surrounding prompt, without relying on additional task-specific or efficient fine-tuning.

\subsection{Vision-Language Models}
Vision-Language Models leverage large-scale image-text pairs collected from the Internet to learn rich cross-modal correlations, achieving remarkable results in various tasks.
Contrastive Language-Image Pretraining (CLIP) \cite{CLIP} employs contrastive learning to align feature representations of vision and language within a joint embedding space, demonstrating impressive zero-shot image classification performance.
Several studies leverage transfer learning \cite{coop,cocoop} or knowledge distillation \cite{distillation1} to extract and adapt knowledge from CLIP, demonstrating superior performance across various downstream tasks, \textit{e.g.} object detection, semantic segmentation, and visual question answering, \textit{etc.}.
Furthermore, with the use of larger training datasets \cite{ALIGN}, varied model designs \cite{stt}, and adaptations for different downstream tasks \cite{Alpha-CLIP,blip2}, many Vision-Language Models have been proposed.
In our work, leveraging the abundant cross-modal knowledge stored in pre-trained Vision-Language Models, we extract instance-level visual features and textual features to achieve cross-modal semantic alignment.

\section{Methodology}

\begin{figure*}[!t]
\centering
\includegraphics[width=\linewidth]{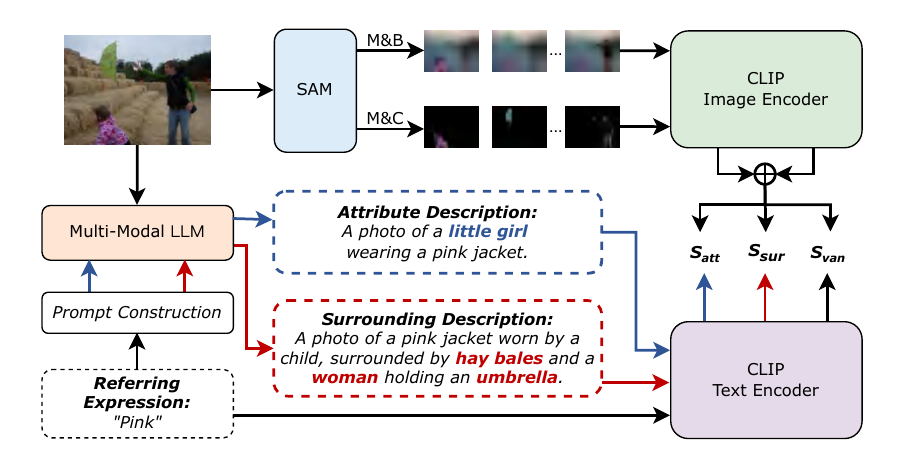}
\caption{{\bf The pipeline of LGD.} We construct the attribute prompt and surrounding prompt by combining instructions and referring expressions. Given the input image and the different prompts, the MLLMs generates the attribute description and surrounding description. After extracting features from the image and text using CLIP, we calculate three visual-text matching scores and obtain the most relevant mask by linearly combining these scores.}
\label{fig_framework}
\end{figure*}


In this section, we detail the proposed LGD, which is straightforward and illustrated in Fig.\ref{fig_framework}.
First, the referring expression and instructions are used to construct the prompts (Section~\ref{Prompt Construction}).
Then, the input image and the two prompts are processed by a MLLMs to generate attribute and surrounding descriptions, and the frozen CLIP Text Encoder extracts feature representations for each description and the original referring expression.
Meanwhile, the off-the-shelf SAM~\cite{SAM} generates instance-level masks from the input image, and these masks are further refined using two mask strategies. The processed masks are then fed into the frozen CLIP Image Encoder to extract the corresponding visual features (Section~\ref{Feature Extraction}).
Next, we introduce three visual-text matching scores, each designed to quantify the alignment between instance-level masks and textual descriptions (Section~\ref{Visual-Text Matching Score}). 
Finally, a linear combination of the three scores is used to select the most accurate mask (Section~\ref{Mask Selection}).

\subsection{Prompt Construction}
\label{Prompt Construction}
Prompt construction is crucial for guiding the MLLMs to produce task-specific outputs.
We design two tailored prompts, $P_{att}$ and $P_{sur}$, to guide the MLLMs in generating the attribute description and the surrounding description.
Specifically, for the crucial attribute description of the referent object, the $P_{att}$ is as follows:

\textit{ Given an image and the corresponding referring expression ``$<input>$'', the entity referred by the referring expression is unique in the image. Please generate a caption with local concept to describe the referent object according to the referring expression. The format is ``A photo of $<object>$ $(attribute)$''.}

Meanwhile, to distinguish the referent object from surrounding objects in the scene, we construct $P_{sur}$, which guides the MLLMs to generate a relevant description. The $P_{sur}$ is as follows:

\textit{Given an image and the corresponding referring expression ``$<input>$'', the entity referred by the referring expression is unique in the image. Please generate a caption to describe the referent object and its surrounding entities according to the referring expression. The format is ``A photo of $<object>$ surrounded by $(entities)$''.}

In $P_{att}$ and $P_{sur}$, $<input>$ refers to the referring expression $T_{van}$, $<object>$ represents the referent object, and $(attribute)$ and $(entities)$ denote the associated descriptions. Notably, $<object>$, $(attribute)$, and $(entities)$ are generated through reasoning by the MLLMs.
To align with the training paradigm of CLIP, we ensure that the generated descriptions begin with ``A photo of''.

\subsection{Feature Extraction}
\label{Feature Extraction}
In this part, we utilize CLIP to extract features of text and image. The text includes the attribute description $T_{att}$, the surrounding description $T_{sur}$, and the referring expression $T_{van}$, and the image includes a set of segmentation proposals generated by an off-the-shelf mask generator.

{\bf{Text Feature Extraction.}} 
We feed the $P_{att}$, $P_{sur}$ and image into MLLMs (\textit{e.g.}, LLaVA \cite{llava}), and the attribute description $T_{att}$ and the surrounding description $T_{sur}$ will be generated through the MLLMs.
The CLIP text encoder $E_t$ is used to obtain textual features $F_{att}^T$ and $F_{van}^T$ as follows:
\begin{align}
F_{att}^T &= E_t(T_{att}), \\
F_{van}^T &= E_t(T_{van}).
\end{align}

For the surrounding description $T_{sur}$, some objects that are not mentioned in the referring expression but identified through $T_{sur}$ contribute significantly to distinguishing the referent object.
To address the issue where the complexity of surrounding description $T_{sur}$ obscures other objects irrelevant to the referent object, we employ Spacy \cite{spacy} to extract noun phrases from $T_{sur}$ and use CLIP to identify those semantically irrelevant to the referent object as negative samples.
We denote the negative samples identified in $T_{sur}$ as $\mathbb{T}_{sur} = \{ t_{sur} \}$.
Following the same approach, the CLIP text encoder $E_t$ is employed to extract textual features $F_{sur}^t$, defined as follows:
\begin{equation}
   F_{sur}^t = E_t(t_{sur}).
\end{equation}

{\bf{Image Feature Extraction.}}
For image $I \in \mathbb{R}^{H \times W \times 3}$, we extract visual features following the BMS \cite{BMS}.
First, an off-the-shelf mask generator (\textit{e.g.}, SAM \cite{SAM}) is employed to produce a set of segmentation proposals $m \in \mathbb{M}$.
Next, for each mask proposal $m$, two mask strategies, $M\&B$ \cite{TAS2023} and $M\&C$ \cite{GLCLIP2023}, are applied to generate distinct instance-level images, $I_{m}^{MB}$ and $I_{m}^{MC}$.
Next, the CLIP visual encoder $E_i$ is utilized to extract instance-level visual features from these images.
\begin{align}
F_{m}^{MB} &= E_i(I_{m}^{MB}), \\
F_{m}^{MC} &= E_i(I_{m}^{MC}).
\end{align}

Finally, $F_{m}^{MB}$ and $F_{m}^{MC}$ are merged to produce the final instance-level visual features $F_{m}^{I}$ as follows:
\begin{equation}
    F_{m}^{I} = F_{m}^{MB} + F_{m}^{MC}.
\end{equation}

\subsection{Visual-Text Matching Score}
\label{Visual-Text Matching Score}
We propose three Visual-Text Matching Scores to thoroughly assess the correspondence between instance-level visual representations and textual representations associated with the referent object.

{\bf{First,}} $S_{att}$ is introduced to quantify the similarity between the instance-level visual features $F_{m}^{I}$ and the crucial attribute textual features $F_{att}^T$, and is calculated as follows:
\begin{equation}
    S_{att} = cos(F_{m}^{I}, F_{att}^T),
\end{equation}
where $cos(,)$ means the cosine similarity between visual and textual features.

{\bf{Second,}} $S_{sur}$ is introduced to quantify the similarity between the instance-level visual features $F_{m}^{I}$ and the textual features of negative samples $F_{sur}^t$ in the surrounding description $T_{sur}$. The calculation involving negative samples and each instance-level visual feature is as follows:
\begin{equation}
     S_{sur} = -\frac{1}{\left| \mathbb{T}_{sur}\right|}\sum cos(F_{m}^{I}, F_{sur}^t),
\end{equation}
where $\left| \mathbb{T}_{sur}\right|$ represents the number of negative samples. Since $S_{sur}$ is designed to quantify the similarity between an instance-level visual feature and textual feature of objects irrelevant to the referent object, it serves as a negative score to identify the least similar mask proposal~\cite{TAS2023,BMS}.

{\bf{Third,}} $S_{van}$ is introduced to quantify the similarity between the instance-level visual features $F_{m}^{I}$ and the vanilla referring expression textual features $F_{van}^T$, and is calculated as follows:
\begin{equation}
    S_{van} = cos(F_{m}^{I}, F_{van}^T).
\end{equation}

\subsection{Mask Selection}
\label{Mask Selection}
A linear combination of the three scores is employed to select the most accurate mask proposal. The final similarity $S$ calculation is as follows:
\begin{equation}
    S = S_{van} + \alpha S_{att} + \beta S_{sur}.
\end{equation}
where $\alpha$ and $\beta$ represent adaptive hyberparameters, respectively.
Therefore, the mask proposal with the highest score is selected as the final output mask $M$:
\begin{equation}
    M = \mathop{\arg\max}S.
\end{equation}

\section{Experiment}

\subsection{Setup}
{\bf{Datasets.}} 
To validate the effectiveness of LGD, we conducted extensive experiments on three widely used referring image segmentation datasets: RefCOCO \cite{refcoco}, RefCOCO+ \cite{refcoco}, and RefCOCOg \cite{refcocog1, refcocog2}. These datasets are standard benchmarks in RIS tasks.
RefCOCO contains 19,994 images and 142,210 referring expressions, emphasizing simple descriptions related to the scene and the location of the referent object;
RefCOCO+ includes 19,992 images and 141,564 referring expressions, focusing on fine-grained visual attributes of the referent object;
RefCOCOg comprises 26,711 images and 104,560 referring expressions, characterized by complex linguistic structures that require models to understand relationships between multiple objects.

{\bf{Evaluation Metrics.}}
Consistent with the evaluation metrics used in existing Zero-shot RIS \cite{GLCLIP2023,TAS2023,BMS}, we adopt overall Intersection over Union (oIoU) and mean Intersection over Union (mIoU) to assess model performance.
oIoU measures the ratio of the total intersection area to the total union area. It evaluates the overall segmentation performance of the model but is sensitive to the presence of large objects.
mIoU calculates the average IoU across all samples, providing a more balanced evaluation of the segmentation results corresponding to each referring expression.

{\bf{Baselines.}}
We evaluate LGD in comparison with several representative baselines, including activation map-based methods (\textit{e.g.}, Grad-CAM \cite{Grad-CAM}, Score Map \cite{ScoreMap} and CLIP-Surgery \cite{ClipSurgery}), Region Token \cite{zrec1}, Global-Local CLIP \cite{GLCLIP2023} and its variant Cropping, TAS \cite{TAS2023}, and CaR \cite{clipasrnn}.
Additionally, some methods (\textit{e.g.}, S+S \cite{ssc-wris} and BMS \cite{BMS}) employ an open-set object detector (\textit{e.g.}, Grounding DINO \cite{Groundingdino}) to generate segmentation prompts, which are subsequently processed by a mask proposal network (\textit{e.g.}, SAM \cite{SAM}) to obtain mask proposals from the input image.
For a fair comparison, we report the results of our method using only SAM as well as the results using the combination of Grounding DINO and SAM.

{\bf{Implementation Details.}}
Our approach works without requiring any training procedure, all experiments are conducted on just one NVIDIA 4090 GPU.
We use LLaVA-v1.6-Vicuna-7B~\cite{liu2024improved} as our MLLMs, where SAM as our mask proposal network.
For CLIP, we report results using ViT-B/32 as backbones, with the same settings as in previous work~\cite{GLCLIP2023,TAS2023}.
For the open-set object detector, we use Grounding DINO with Swin-T.
We set $\beta = 1$ for all datasets, $\alpha = 0.5$ for RefCOCO and RefCOCO+, and $\alpha = 0.3$ for RefCOCOg.

\subsection{Comparison with State-of-the-Art Methods}

\begin{table*}[ht]
\caption{Comparison with Zero-shot RIS methods on three standard benchmark datasets. U refers to the UMD partition, and G refers to the Google partition. The best results are highlighted in bold, while the second-best results are underlined. These results demonstrate the superior performance of our model. \label{tab:mainresult}}
\centering
\resizebox{\textwidth}{!}{
\begin{tabular}{c|c|l|c|ccc|ccc|ccc}
\hline
\multirow{2}{*}{Metric} &
  \multirow{2}{*}{Methods} &
  \multirow{2}{*}{Backbones} &
  \multirow{2}{*}{Venue} &
  \multicolumn{3}{c|}{RefCOCO} &
  \multicolumn{3}{c|}{RefCOCO+} &
  \multicolumn{3}{c}{RefCOCOg} \\
 &
   &
   &
   &
  Val &
  TestA &
  TestB &
  Val &
  TestA &
  TestB &
  Val(U) &
  Test(U) &
  Test(G) \\ \hline
\multirow{12}{*}{oIoU} &
  Grad-CAM \cite{Grad-CAM} &
  ResNet-50 &
  IJCV20 &
  23.44 &
  23.91 &
  21.60 &
  26.67 &
  27.20 &
  24.84 &
  23.00 &
  23.91 &
  23.57 \\
 &
  Score Map \cite{ScoreMap} &
  ResNet-50 &
  ECCV22 &
  20.18 &
  20.52 &
  21.30 &
  22.06 &
  22.43 &
  24.61 &
  23.05 &
  23.41 &
  23.69 \\
 &
  CLIP-Surgery \cite{ClipSurgery} &
  ResNet-50 &
  CVPR23 &
  18.04 &
  14.74 &
  21.28 &
  18.39 &
  14.34 &
  22.98 &
  20.44 &
  21.80 &
  21.23 \\
 &
  Region Token \cite{zrec1} &
  ViT-B/32 &
  arXiv &
  11.56 &
  12.37 &
  11.35 &
  12.54 &
  14.07 &
  12.22 &
  10.87 &
  11.51 &
  11.74 \\
 &
  Cropping \cite{GLCLIP2023} &
  ViT-B/32 &
  CVPR23 &
  22.73 &
  21.11 &
  23.08 &
  24.09 &
  22.42 &
  23.93 &
  28.69 &
  27.51 &
  27.70 \\
 &
  Global-Local CLIP \cite{GLCLIP2023} &
  ViT-B/32 &
  CVPR23 &
  24.88 &
  23.61 &
  24.66 &
  26.16 &
  24.90 &
  25.83 &
  31.11 &
  30.96 &
  30.69 \\
 &
  Global-Local CLIP (SAM) \cite{GLCLIP2023} &
  ViT-B/32 &
  CVPR23 &
  22.43 &
  24.66 &
  21.27 &
  26.35 &
  30.80 &
  22.65 &
  27.57 &
  27.87 &
  27.80 \\
 &
  TAS \cite{TAS2023} &
  ViT-B/32 &
  ACL23 &
  {\underline{29.53}} &
  {\underline{30.26}} &
  {\underline{28.24}} &
  {\underline{33.21}} &
  {\underline{38.77}} &
  {\underline{28.01}} &
  {\underline{35.84}} &
  {\underline{36.16}} &
  {\underline{36.36}} \\
 &
  LGD &
  ViT-B/32 &
  - &
  \textbf{34.47} &
  \textbf{37.81} &
  \textbf{30.79} &
  \textbf{36.72} &
  \textbf{44.36} &
  \textbf{30.11} &
  \textbf{36.57} &
  \textbf{37.63} &
  \textbf{38.19} \\ \cline{2-13} 
 &
  S+S \cite{ssc-wris} &
  ViT-B/32 &
  arXiv &
  33.31 &
  40.35 &
  26.14 &
  34.84 &
  43.16 &
  28.22 &
  35.71 &
  42.10 &
  41.70 \\
 &
  BMS \cite{BMS} &
  \multicolumn{1}{c|}{-} &
  TCSVT24 &
  {\underline{34.94}} &
  {\underline{43.34}} &
  {\underline{26.27}} &
  {\underline{37.36}} &
  {\underline{46.96}} &
  {\underline{28.57}} &
  {\underline{42.61}} &
  {\underline{44.62}} &
  {\underline{44.84}} \\
 &
  LGD+DINO &
  ViT-B/32 &
  - &
  \textbf{42.72} &
  \textbf{48.20} &
  \textbf{36.24} &
  \textbf{47.26} &
  \textbf{50.93} &
  \textbf{34.43} &
  \textbf{45.54} &
  \textbf{47.65} &
  \textbf{49.37} \\ \hline
\multirow{13}{*}{mIoU} &
  Grad-CAM \cite{Grad-CAM} &
  ResNet-50 &
  IJCV20 &
  30.22 &
  31.90 &
  27.17 &
  33.96 &
  25.66 &
  32.29 &
  33.05 &
  32.50 &
  33.25 \\
 &
  Score Map \cite{ScoreMap} &
  ResNet-50 &
  ECCV22 &
  25.62 &
  26.66 &
  25.17 &
  27.49 &
  28.49 &
  30.47 &
  30.13 &
  30.15 &
  31.10 \\
 &
  CLIP-Surgery \cite{ClipSurgery} &
  ResNet-50 &
  CVPR23 &
  25.03 &
  22.71 &
  27.12 &
  26.01 &
  22.20 &
  30.18 &
  29.26 &
  30.07 &
  29.43 \\
 &
  Region Token \cite{zrec1} &
  ViT-B/32 &
  arXiv &
  17.06 &
  18.02 &
  16.28 &
  18.83 &
  20.31 &
  17.78 &
  16.33 &
  16.88 &
  17.31 \\
 &
  Cropping \cite{GLCLIP2023} &
  ViT-B/32 &
  CVPR23 &
  24.83 &
  22.58 &
  25.72 &
  26.33 &
  24.06 &
  26.46 &
  31.88 &
  30.94 &
  31.06 \\
 &
  Global-Local CLIP \cite{GLCLIP2023} &
  ViT-B/32 &
  CVPR23 &
  26.20 &
  24.94 &
  26.56 &
  27.80 &
  25.64 &
  27.84 &
  33.52 &
  33.67 &
  33.61 \\
 &
  Global-Local CLIP (SAM) \cite{GLCLIP2023} &
  ViT-B/32 &
  CVPR23 &
  32.93 &
  34.93 &
  30.09 &
  38.37 &
  42.05 &
  32.65 &
  42.02 &
  42.02 &
  42.67 \\
 &
  TAS \cite{TAS2023} &
  ViT-B/32 &
  ACL23 &
  {\underline{39.84}} &
  {\underline{41.08}} &
  {\underline{36.24}} &
  {\underline{43.63}} &
  {\underline{49.13}} &
  {\underline{36.54}} &
  {\underline{46.62}} &
  {\underline{46.80}} &
  {\underline{48.05}} \\
 &
  CaR \cite{clipasrnn} &
  ViT-B/16 &
  CVPR24 &
  33.57 &
  35.36 &
  30.51 &
  34.22 &
  36.03 &
  31.02 &
  36.67 &
  36.57 &
  36.63 \\
 &
  LGD &
  ViT-B/32 &
  - &
  \textbf{45.41} &
  \textbf{48.81} &
  \textbf{39.28} &
  \textbf{47.23} &
  \textbf{54.64} &
  \textbf{38.10} &
  \textbf{47.31} &
  \textbf{48.18} &
  \textbf{49.31} \\ \cline{2-13} 
 &
  S+S \cite{ssc-wris} &
  ViT-B/32 &
  arXiv &
  36.95 &
  43.77 &
  27.97 &
  37.68 &
  46.24 &
  29.31 &
  41.41 &
  47.18 &
  47.57 \\
 &
  BMS \cite{BMS} &
  \multicolumn{1}{c|}{-} &
  TCSVT24 &
  {\underline{40.77}} &
  {\underline{49.64}} &
  {\underline{29.66}} &
  {\underline{41.70}} &
  {\underline{51.18}} &
  {\underline{30.16}} &
  {\underline{48.32}} &
  {\underline{50.16}} &
  {\underline{50.26}} \\
 &
  LGD+DINO &
  ViT-B/32 &
  - &
  \textbf{49.54} &
  \textbf{54.67} &
  \textbf{40.95} &
  \textbf{49.57} &
  \textbf{58.36} &
  \textbf{38.55} &
  \textbf{50.29} &
  \textbf{51.09} &
  \textbf{52.51} \\ \hline
\end{tabular}
}
\end{table*}

In this section, we compare LGD with recent state-of-the-art approaches on three benchmarks in terms of oIoU and mIoU metrics.
As shown in Table~\ref{tab:mainresult}, our method outperforms all the compared methods.
Specifically, compared to the current state-of-the-art method TAS, our approach achieves oIoU improvements of 4.94\%, 7.55\%, and 2.55\% on RefCOCO; 3.51\%, 5.59\%, and 2.1\% on RefCOCO+; and 0.73\%, 1.47\%, and 1.83\% on RefCOCOg.
Meanwhile, our approach achieves mIoU improvements of 5.57\%, 7.73\%, and 3.04\% on RefCOCO; 3.6\%, 5.51\%, and 1.56\% on RefCOCO+; and 0.69\%, 1.38\%, and 1.26\% on RefCOCOg.
The consistent oIoU and mIoU improvements over TAS \cite{TAS2023} validate the superiority of our approach in handling diverse application scenario. 
On RefCOCO, which emphasizes simple scene and location descriptions, our model achieves notable average gains of 5.01\% in oIoU and 5.45\% in mIoU, demonstrating its capability to segment referents described by relatively straightforward referring expressions.
On RefCOCO+, our method achieves average improvements of 3.73\% in oIoU and 3.56\% in mIoU, highlighting its ability to capture fine-grained visual distinctions.
On RefCOCOg, which involves complex linguistic structures and relationships among multiple objects, our method still achieves competitive gains, with average improvements of 1.34\% in oIoU and 1.11\% in mIoU.
Most baseline methods focus on the global features of the referent object and fail to capture crucial attribute details or effectively distinguish the referent object from its surrounding objects, leading to suboptimal performance.
The competitive performance of LGD stems from the incorporation of attribute description and surrounding description.
Attribute description provides a fine-grained characterization of the referent object, helping to reduce the ambiguity and diversity in referring expressions.
Meanwhile, surrounding description adds valuable contextual information by explicitly distinguishing the referent object from its surrounding objects.
These descriptions effectively enhance the interpretation of referring expressions, allowing the model to accurately locate the referent object.

In addition, for a fair comparison with S+S~\cite{ssc-wris} and BMS~\cite{BMS}, which utilize DINO~\cite{Groundingdino} to generate visual prompts (\textit{i.e.}, box prompts) for SAM, we report the results of our method using the same settings as LGD+DINO in Table~\ref{tab:mainresult}.
Specifically, compared to the current state-of-the-art method BMS, our approach achieves oIoU improvements of 7.78\%, 4.86\%, and 9.97\% on RefCOCO; 9.90\%, 3.97\%, and 5.86\% on RefCOCO+; and 2.93\%, 3.03\%, and 4.53\% on RefCOCOg.
Meanwhile, our approach achieves mIoU improvements of 8.77\%, 5.03\%, and 11.29\% on RefCOCO; 7.87\%, 7.18\%, and 8.39\% on RefCOCO+; and 1.97\%, 0.93\%, and 2.25\% on RefCOCOg.
The precise box prompts generated by DINO provide a more reliable foundation for SAM, improving the quality and accuracy of mask proposals. These improved mask proposals enable LGD to achieve substantial performance gains across all datasets.
This demonstrates that combining open-set detectors (\textit{i.e.}, DINO) with segmentation models (\textit{i.e.}, SAM) can significantly enhance referring image segmentation, particularly for tasks requiring fine-grained segmentation and understanding of complex object relationships.

\subsection{Ablation Study}

\begin{table*}[ht]
\caption{Effects of Visual-Text Matching Score on three standard benchmark datasets. U refers to the UMD partition, and G refers to the Google partition. The best results are highlighted in bold. \label{tab:MatScore}}
\centering
\resizebox{\textwidth}{!}{
\begin{tabular}{c|ccc|ccc|ccc|ccc}
\hline
\multirow{2}{*}{Metric} &
  \multicolumn{3}{c|}{Visual-Text Matching Score} &
  \multicolumn{3}{c|}{RefCOCO} &
  \multicolumn{3}{c|}{RefCOCO+} &
  \multicolumn{3}{c}{RefCOCOg} \\ \cline{2-13} 
                      & $S_{van}$  & $S_{att}$  & $S_{sur}$  & Val   & TestA & TestB & Val   & TestA & TestB & Val(U) & Test(U) & Test(G) \\ \hline
\multirow{4}{*}{oIoU} & \checkmark &            &            & 26.26 & 27.19 & 27.70 & 31.01 & 36.27 & 27.15 & 32.89  & 34.03   & 33.42   \\
                      & \checkmark &            & \checkmark & 29.36 & 30.29 & 29.12 & 33.50  & 39.65 & 28.45 & 35.32  & 37.24   & 36.61   \\
                      & \checkmark & \checkmark &            & 30.69 & 33.34 & 28.71 & 33.36 & 39.16 & 27.93 & 32.49  & 33.43   & 33.30   \\
 &
  \checkmark &
  \checkmark &
  \checkmark &
  \textbf{34.47} &
  \textbf{37.81} &
  \textbf{30.79} &
  \textbf{36.72} &
  \textbf{44.36} &
  \textbf{30.11} &
  \textbf{36.57} &
  \textbf{37.63} &
  \textbf{38.19} \\ \hline
\multirow{4}{*}{mIoU} & \checkmark &            &            & 37.77 & 38.70  & 36.06 & 42.28 & 46.94 & 35.33 & 44.20  & 44.78   & 45.50   \\
                      & \checkmark &            & \checkmark & 40.21 & 41.25 & 36.55 & 43.67 & 49.65 & 36.10 & 46.12  & 47.31   & 47.54   \\
                      & \checkmark & \checkmark &            & 42.25 & 45.24 & 38.09 & 44.68 & 50.69 & 36.31 & 43.91  & 44.56   & 45.52   \\
 &
  \checkmark &
  \checkmark &
  \checkmark &
  \textbf{45.41} &
  \textbf{48.81} &
  \textbf{39.28} &
  \textbf{47.23} &
  \textbf{54.64} &
  \textbf{38.10} &
  \textbf{47.31} &
  \textbf{48.18} &
  \textbf{49.31} \\ \hline
\end{tabular}%
}
\end{table*}

{\bf{Effects of Visual-Text Matching Score.}}
To evaluate the contribution of different components of our visual-text matching score, we conduct ablation studies with four configurations: 
$S_{van}$, $S_{van} + S_{sur}$, $S_{van} + S_{att}$, and $S_{van} + S_{att} + S_{sur}$.
We evaluate these setups on RefCOCO , RefCOCO+ and RefCOCOg, reporting both oIoU and mIoU. The experimental results are summarized in Table \ref{tab:MatScore}.
The $S_{van}$ achieves reasonable performance by relying solely on instance-level visual-text similarity. However, the absence of contextual information limits its ability to handle ambiguous or complex referring expressions, leading to lower scores across all metrics.
$S_{van} + S_{sur}$ significantly improves performance, showing the importance of using surrounding object descriptions to clarify the referent object.
In RefCOCO and RefCOCO+, $S_{van} + S_{att}$ significantly improves compared to $S_{van}$, demonstrating that attribute description help capture a more comprehensive and detailed characterization of the referent object. However, performance on RefCOCOg declines. We believe this is because the referring expressions in RefCOCOg already provide sufficient key attributes, making the additional attribute description less beneficial.
$S_{van} + S_{att} + S_{sur}$ achieves the best performance across all splits.
The design of the three Visual-Text Matching Scores provides distinct advantages.
$S_{att}$ emphasizes the correspondence between instance-level visual features and the crucial attributes of the referent object.
$S_{sur}$ explicitly distinguishes the referent object from its surrounding objects, providing additional spatial and contextual information that significantly enhances localization.
$S_{van}$ focuses on aligning the instance-level visual features of the referent object with the referring expression.
These scores complement each other to form a comprehensive visual-text matching strategy, facilitating a more holistic understanding of the referent object.


\begin{figure}[thb] 
    \centering
    \includegraphics[width=0.32\textwidth]{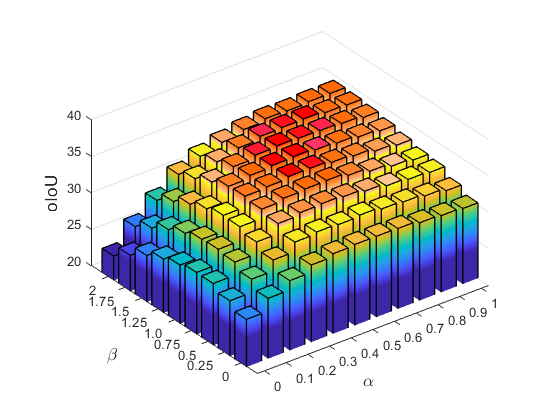}
    \includegraphics[width=0.32\textwidth]{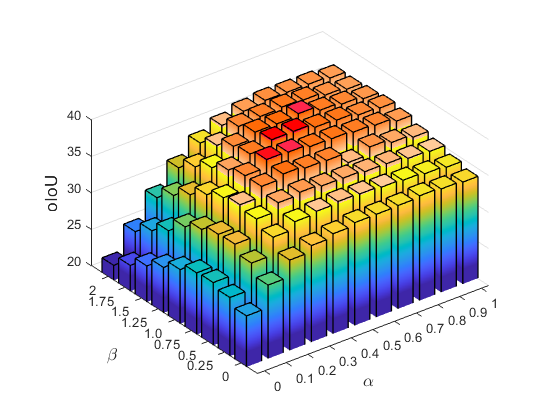} 
    \includegraphics[width=0.32\textwidth]{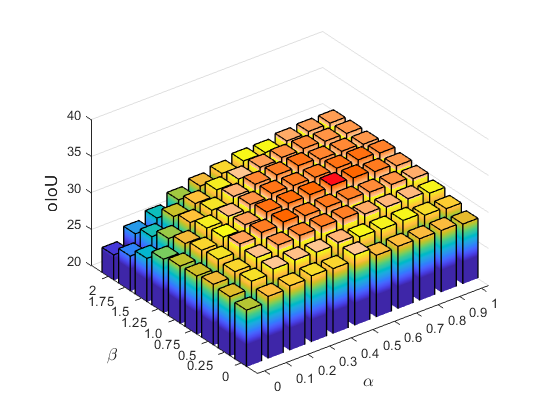}
    \\
    \makebox[0.32\textwidth]{\scriptsize (a) oIoU on RefCOCO Val}
    \makebox[0.32\textwidth]{\scriptsize (b) oIoU on RefCOCO TestA}
    \makebox[0.32\textwidth]{\scriptsize (c) oIoU on RefCOCO TestB}
    \\
    \includegraphics[width=0.32\textwidth]{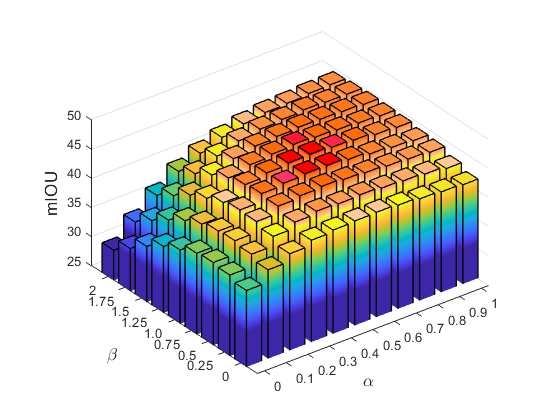}
    \includegraphics[width=0.32\textwidth]{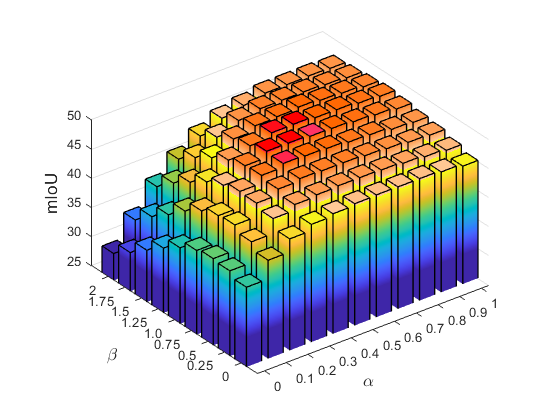}
    \includegraphics[width=0.32\textwidth]{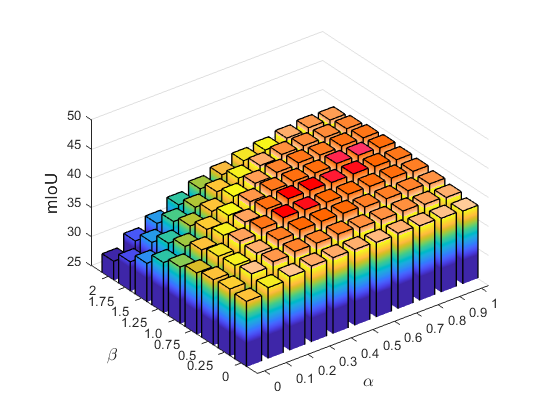}
    \\
    \makebox[0.32\textwidth]{\scriptsize (d) mIoU on RefCOCO Val}
    \makebox[0.32\textwidth]{\scriptsize (e) mIoU on RefCOCO TestA}
    \makebox[0.32\textwidth]{\scriptsize (f) mIoU on RefCOCO TestB}
    \caption{{\bf Sensitive toward $\alpha$ and $\beta$ on RefCOCO.}} 
    \label{fig_sen}
\end{figure}

\begin{figure*} 
    \centering
    \includegraphics[width=0.32\textwidth]{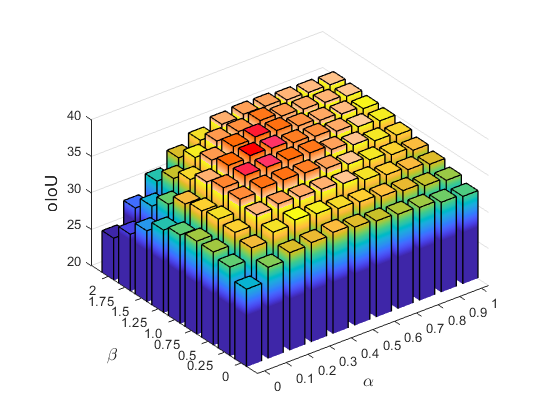}
    \includegraphics[width=0.32\textwidth]{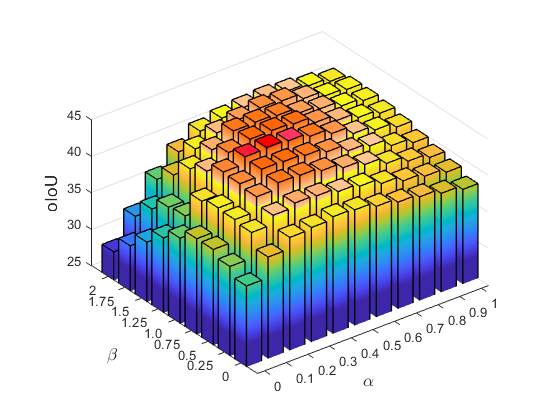} 
    \includegraphics[width=0.32\textwidth]{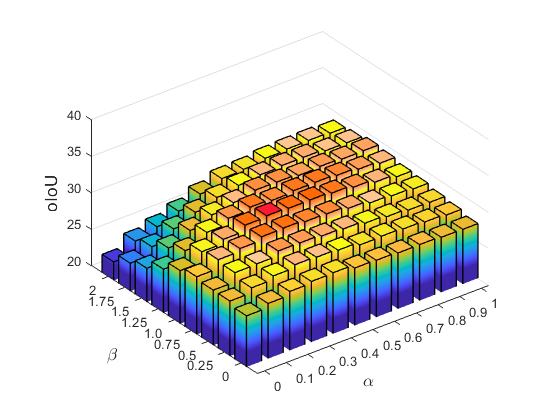}
    \\
    \makebox[0.32\textwidth]{\tiny (a) oIoU on RefCOCO+ Val}
    \makebox[0.32\textwidth]{\tiny (b) oIoU on RefCOCO+ TestA}
    \makebox[0.32\textwidth]{\tiny (c) oIoU on RefCOCO+ TestB}
    \\
    \includegraphics[width=0.32\textwidth]{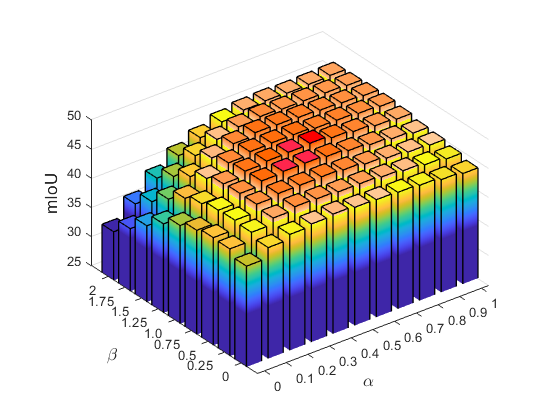}
    \includegraphics[width=0.32\textwidth]{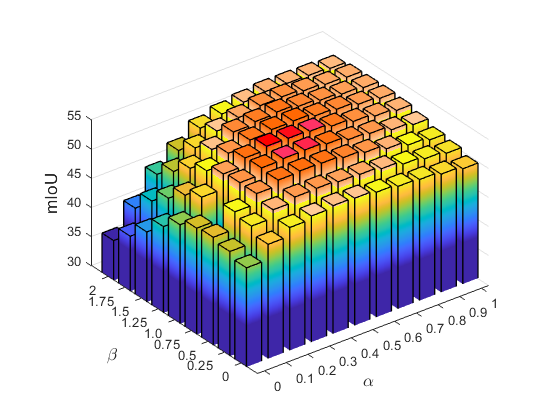}
    \includegraphics[width=0.32\textwidth]{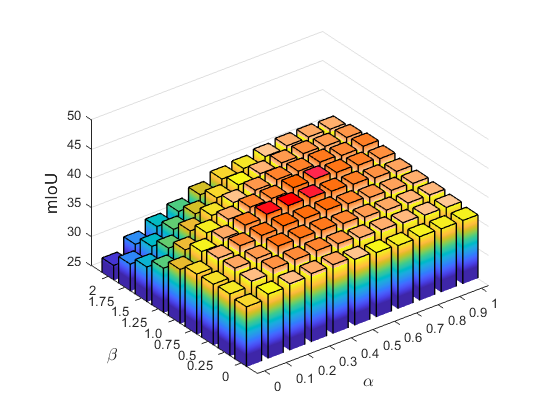}
    \\
    \makebox[0.32\textwidth]{\tiny (d) mIoU on RefCOCO+ Val}
    \makebox[0.32\textwidth]{\tiny (e) mIoU on RefCOCO+ TestA}
    \makebox[0.32\textwidth]{\tiny (f) mIoU on RefCOCO+ TestB}
    \caption{{\bf Sensitive toward $\alpha$ and $\beta$ on RefCOCO+.}} 
    \label{fig_sen_+}
\end{figure*}

\begin{figure*}
    \centering
    \includegraphics[width=0.32\textwidth]{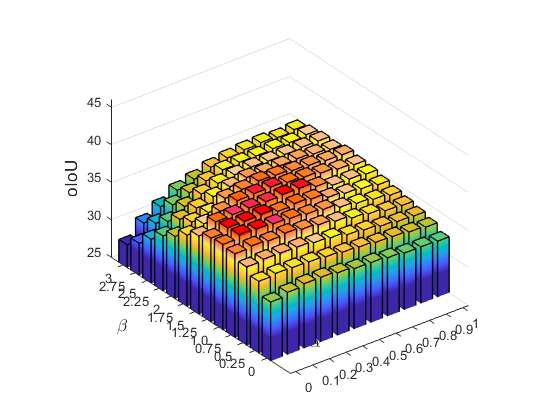}
    \includegraphics[width=0.32\textwidth]{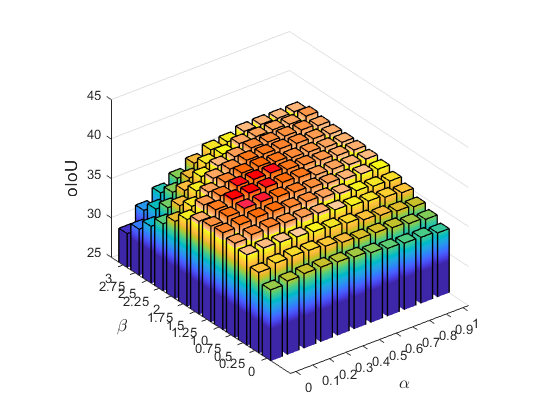} 
    \includegraphics[width=0.32\textwidth]{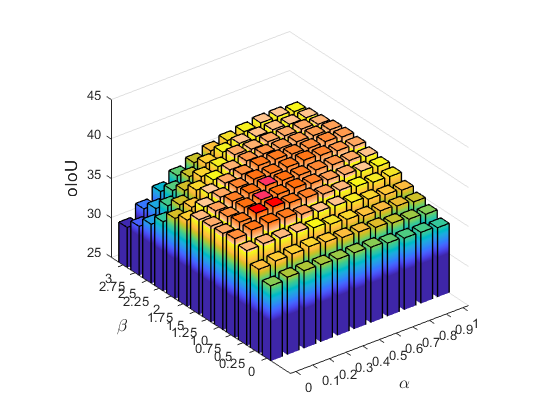}
    \\
    \makebox[0.32\textwidth]{\tiny (a) oIoU on RefCOCOg Val(u)}
    \makebox[0.32\textwidth]{\tiny (b) oIoU on RefCOCOg Test(u)}
    \makebox[0.32\textwidth]{\tiny (c) oIoU on RefCOCOg Test(g)}
    \\
    \includegraphics[width=0.32\textwidth]{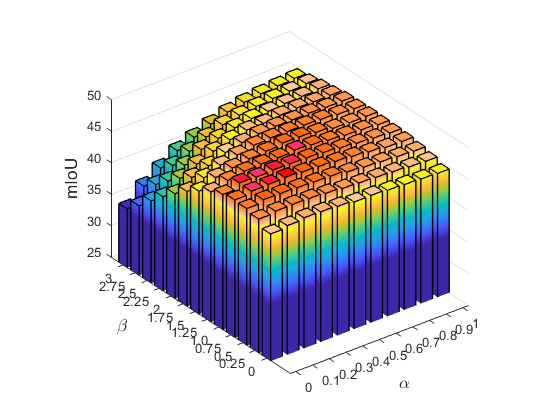}
    \includegraphics[width=0.32\textwidth]{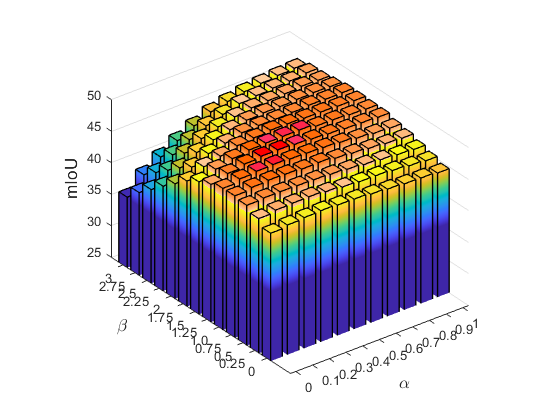}
    \includegraphics[width=0.32\textwidth]{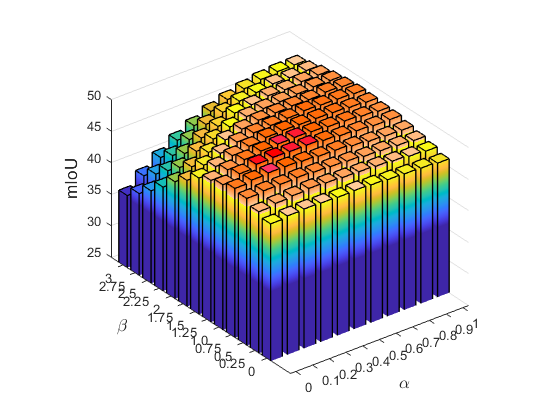}
    \\
    \makebox[0.32\textwidth]{\tiny (d) mIoU on RefCOCOg Val(u)}
    \makebox[0.32\textwidth]{\tiny (e) mIoU on RefCOCOg Test(u)}
    \makebox[0.32\textwidth]{\tiny (f) mIoU on RefCOCOg Test(g)}
    \caption{{\bf Sensitive toward $\alpha$ and $\beta$ on RefCOCOg.}} 
    \label{fig_sen_g}
\end{figure*}

{\bf{Sensitive toward $\alpha$ and $\beta$.}} 
We conduct a sensitivity analysis with respect to $\alpha$ and $\beta$, where $\alpha$ and $\beta$ control the contributions of $S_{att}$ and $S_{sur}$, respectively.
We first evaluate across various values of $\alpha$ and $\beta$ on the Val, TestA, and TestB of RefCOCO. The results for oIoU and mIoU are reported in Fig.\ref{fig_sen}.

Our experiments show that as $\alpha$ increases, performance improves due to the enhanced contribution of attribute descriptions in capturing comprehensive and detailed characterization of the referent object.
However, when $\alpha$ becomes too large, performance starts to decline, due to an overemphasis on crucial attributes, which causes other important information to be overlooked.
Similarly, increasing $\beta$ demonstrates that surrounding descriptions help distinguish the referent object from irrelevant objects.
However, excessively large values of $\beta$ can introduce noise, leading to performance degradation.
We observe that $\alpha=0.5$ and $\beta=1$ yield an effective combination, demonstrating that the balance between attribute descriptions and surrounding descriptions is crucial for achieving precise object segmentation.




We extend the sensitivity analysis of parameters $\alpha$ and $\beta$ to RefCOCO+ and RefCOCOg, with full results shown in Fig.\ref{fig_sen_+} and Fig.\ref{fig_sen_g}, respectively. Similar to the observations on RefCOCO, both datasets show consistent trends. Increasing $\alpha$ within a moderate range improves performance, as attribute descriptions help characterize the target object. However, when $\alpha > 0.8$, the performance drops, indicating that overemphasizing attributes may ignore other important cues. A similar pattern is observed for $\beta$: raising it improves disambiguation by leveraging surrounding descriptions, but values larger than 1.5 introduce noise and reduce accuracy.

We observe that the optimal $\alpha$ varies across datasets. $\alpha=0.5$ performs best on RefCOCO and RefCOCO+, while a smaller value $\alpha=0.3$ works better on RefCOCOg, possibly because of the increased complexity of its expressions.

These results confirm that combining attribute and surrounding descriptions in a balanced way is essential. Too much focus on either one leads to performance saturation or decline, highlighting the importance of integrating both types of information for accurate segmentation.


\begin{figure}[!t] 
    \centering
    \includegraphics[width=0.45\textwidth]{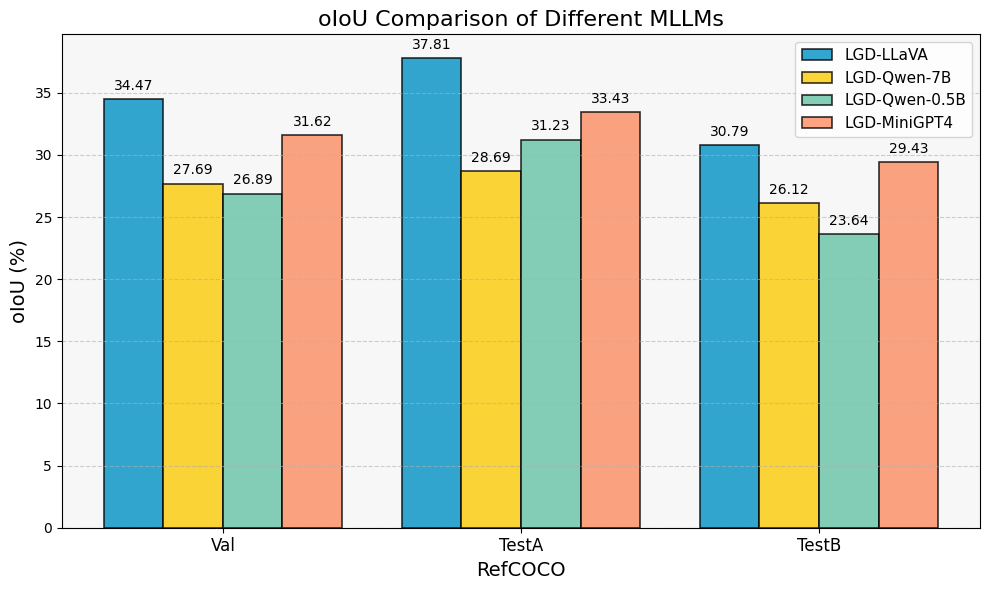}
    \includegraphics[width=0.45\textwidth]{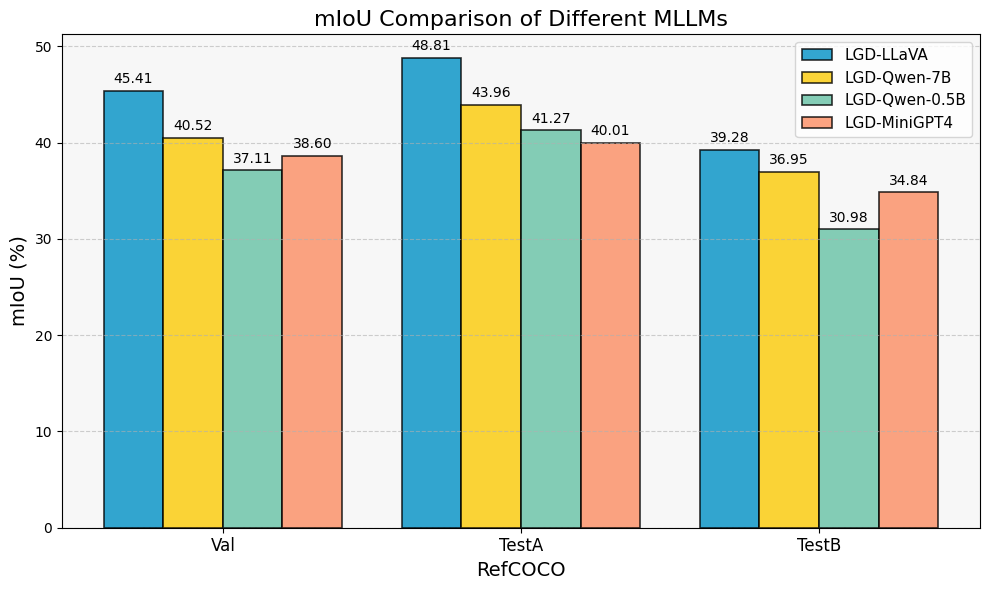}
    \caption{{\bf Effects of Different MLLMs.}}
    \label{fig_MLLMs}
\end{figure}

{\bf{Effects of Different MLLMs.}}
We conduct an ablation study to investigate the effects of different MLLMs on performance.
The study compares four setups: LGD-LLaVa, LGD-Qwen-VL-s7B \cite{bai2023qwen}, LGD-Qwen-VL-0.5B \cite{bai2023qwen},and LGD-MiniGPT-4 \cite{gpt4}.
Results are evaluated on RefCOCO across the Val, TestA, and TestB using oIoU and mIoU metrics in Fig.\ref{fig_MLLMs}.
The experimental results show that LGD-LLaVa achieves the best performance across all splits, demonstrating its ability to effectively use attribute descriptions and surrounding descriptions to improve segmentation accuracy.
In comparison, other MLLMs perform worse in the zero-shot RIS task, likely due to their insufficient ability to generate crucial textual descriptions using multi-modal features.
This limitation may lead to less accurate representation of key attributes and less effective differentiation between referent and irrelevant objects.
The results highlight the importance of selecting an MLLMs with strong cross-modal alignment to achieve precise segmentation.

\subsection{Qualitative Results}

\begin{figure*}[ht]
\centering
\includegraphics[width=\linewidth]{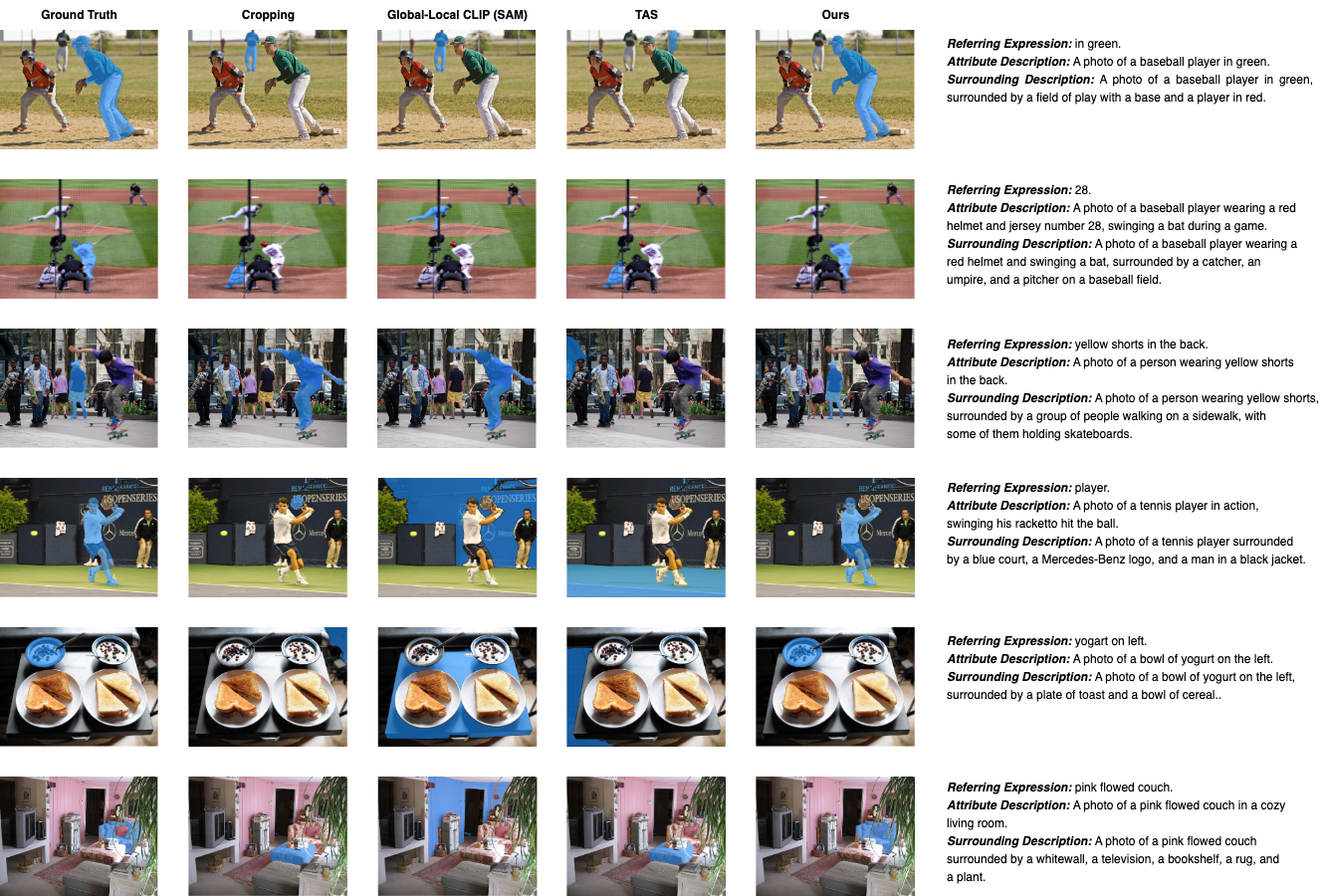}
\caption{{\bf Qualitative results of different methods.} 
From left to right are the Ground Truth and the visualization results of Cropping, Global-Local CLIP (SAM), TAS, and LGD.
The last column presents the referring expression associated with the referent object, along with attribute description and surrounding description generated by the MLLMs.
All referring expressions are drawn from the RefCOCO.
}
\label{vis}
\end{figure*}

We present the visualization results in Fig.~\ref{vis} to compare the performance of our proposed method LGD with several representative baselines, including Cropping \cite{GLCLIP2023}, Global-Local CLIP (SAM) \cite{GLCLIP2023}, and TAS \cite{TAS2023}. The results are obtained using publicly available codes and models of these baseline methods.
The experimental results demonstrate the effectiveness of our method in addressing the ambiguity and diversity of free-form referring expressions while mitigating interference from complex scenes.
The introduction of attribute description and surrounding description effectively captures fine-grained features of referent objects while distinguishing them from irrelevant objects in the visual scene.
Specifically, when a referring expression implicitly describes the referent object with description like \textit{``in green''}, \textit{``28''}, or \textit{``yellow short in the back''}, the inherent ambiguity and diversity of such expressions challenge the cross-modal alignment capabilities of CLIP. As a result, baseline methods struggle to accurately locate the referring object.
In contrast, the attribute prompt we introduced guides the MLLMs to generate attribute description, enabling it to capture the entity category and key attributes of the target object, effectively addressing these challenges.
Furthermore, distinguishing between referent objects and irrelevant objects in the visual scene presents a significant challenge. Referring expressions tend to lack relevant descriptions, such as \textit{``player''}, \textit{``yogart on left''} \footnote{The dataset contains the typo ``yogart''.}, or \textit{pink floral couch''}. We introduce the surrounding prompt to guide MLLMs in generating surrounding description, which provide context for irrelevant objects around the referent object, thereby enhancing LGD to distinguish the referent object.
In summary, the visualization results demonstrate that our method outperforms existing approaches, achieving more accurate and reliable segmentation in complex scenarios.

\section{Conclusion}
In this paper, we proposed LGD, a framework designed to enhance the region-text matching ability of Vision-Language Models using specific descriptions generated by MLLMs.
To achieve this, we introduced two key prompts: the attribute prompt and the surrounding prompt. These prompts guide MLLMs to generate attribute description for the crucial attributes of the referring object and surrounding description for objects not related to the referring expression.
We also designed three visual-text matching scores, leveraging Vision-Language Models to calculate the similarity between instance-level visual and textual features, enabling the identification of the mask proposal most relevant to the referring expression.
By addressing incorrect target localization caused by the inherent ambiguity and diversity of free-form referring expressions, our method achieves more accurate and reliable segmentation in complex scenarios.
LGD achieves competitive results on public datasets, setting a new state-of-the-art performance.

\section*{Acknowledgement}
This research is partially supported by National Natural Science Foundation of China (Grant No. 62271360) and National Key R\&D Program of China (Grant No. 2024YFF0907000).

\bibliographystyle{elsarticle-num}
\bibliography{elsarticle-template-num.bib}

\begin{thebibliography}{10}
\expandafter\ifx\csname url\endcsname\relax
  \def\url#1{\texttt{#1}}\fi
\expandafter\ifx\csname urlprefix\endcsname\relax\def\urlprefix{URL }\fi
\expandafter\ifx\csname href\endcsname\relax
  \def\href#1#2{#2} \def\path#1{#1}\fi

\bibitem{ss1}
Y.~Zhang, M.~A. Mazurowski, Convolutional neural networks rarely learn shape for semantic segmentation, Pattern Recognition 146 (2024) 110018.

\bibitem{ss2}
J.~Jiang, X.~He, X.~Zhu, W.~Wang, J.~Liu, Cgvit: Cross-image groupvit for zero-shot semantic segmentation, Pattern Recognition 164 (2025) 111505.

\bibitem{ss3}
M.~Liao, W.~Li, C.~Yin, Y.~Jin, Y.~Peng, Concept-guided domain generalization for semantic segmentation, Pattern Recognition 164 (2025) 111550.

\bibitem{ss4}
H.~Zheng, X.~Lin, H.~Liang, B.~Zhou, Y.~Liang, Region-aware mutual relational knowledge distillation for semantic segmentation, Pattern Recognition 161 (2025) 111319.

\bibitem{is1}
M.~Ye-Bin, D.~Choi, Y.~Kwon, J.~Kim, T.-H. Oh, Eninst: Enhancing weakly-supervised low-shot instance segmentation, Pattern Recognition 145 (2024) 109888.

\bibitem{is2}
W.~Ouyang, Z.~Xu, J.~Xu, Q.~Wang, Y.~Xu, Mixingmask: A contour-aware approach for joint object detection and instance segmentation, Pattern Recognition 155 (2024) 110620.

\bibitem{is3}
C.~Wang, G.~Wang, Q.~Zhang, P.~Guo, W.~Liu, X.~Wang, Openinst: A simple query-based method for open-world instance segmentation, Pattern Recognition 153 (2024) 110570.

\bibitem{ps1}
T.~Chu, W.~Cai, Q.~Liu, Learning panoptic segmentation through feature discriminability, Pattern Recognition 122 (2022) 108240.

\bibitem{ovs1}
F.~Liang, B.~Wu, X.~Dai, K.~Li, Y.~Zhao, H.~Zhang, P.~Zhang, P.~Vajda, D.~Marculescu, Open-vocabulary semantic segmentation with mask-adapted clip, in: 2023 IEEE/CVF Conference on Computer Vision and Pattern Recognition (CVPR), 2023, pp. 7061--7070.

\bibitem{app11}
J.~Chen, Y.~Shen, J.~Gao, J.~Liu, X.~Liu, Language-based image editing with recurrent attentive models, in: 2018 IEEE/CVF Conference on Computer Vision and Pattern Recognition, 2018, pp. 8721--8729.

\bibitem{app21}
J.~Gu, E.~Stefani, Q.~Wu, J.~Thomason, X.~Wang, Vision-and-language navigation: A survey of tasks, methods, and future directions, in: Proceedings of the 60th Annual Meeting of the Association for Computational Linguistics (Volume 1: Long Papers), 2022, pp. 7606--7623.

\bibitem{app31}
K.~Marino, M.~Rastegari, A.~Farhadi, R.~Mottaghi, Ok-vqa: A visual question answering benchmark requiring external knowledge, in: 2019 IEEE/CVF Conference on Computer Vision and Pattern Recognition (CVPR), 2019, pp. 3190--3199.

\bibitem{cris-ris}
Z.~Wang, Y.~Lu, Q.~Li, X.~Tao, Y.~Guo, M.~Gong, T.~Liu, Cris: Clip-driven referring image segmentation, in: 2022 IEEE/CVF Conference on Computer Vision and Pattern Recognition (CVPR), 2022, pp. 11676--11685.

\bibitem{vlt-ris}
H.~Ding, C.~Liu, S.~Wang, X.~Jiang, Vlt: Vision-language transformer and query generation for referring segmentation, IEEE Transactions on Pattern Analysis and Machine Intelligence 45~(6) (2023) 7900--7916.

\bibitem{lavt-ris}
Z.~Yang, J.~Wang, Y.~Tang, K.~Chen, H.~Zhao, P.~H. Torr, Lavt: Language-aware vision transformer for referring image segmentation, in: 2022 IEEE/CVF Conference on Computer Vision and Pattern Recognition (CVPR), 2022, pp. 18134--18144.

\bibitem{groupformer-ris}
J.~Tang, G.~Zheng, C.~Shi, S.~Yang, Contrastive grouping with transformer for referring image segmentation, in: 2023 IEEE/CVF Conference on Computer Vision and Pattern Recognition (CVPR), 2023, pp. 23570--23580.

\bibitem{kim-wris}
D.~Kim, N.~Kim, C.~Lan, S.~Kwak, Shatter and gather: Learning referring image segmentation with text supervision, in: 2023 IEEE/CVF International Conference on Computer Vision (ICCV), 2023, pp. 15501--15511.

\bibitem{lee-wris}
J.~Lee, S.~Lee, J.~Nam, S.~Yu, J.~Do, T.~Taghavi, Weakly supervised referring image segmentation with intra-chunk and inter-chunk consistency, in: 2023 IEEE/CVF International Conference on Computer Vision (ICCV), 2023, pp. 21813--21824.

\bibitem{ppt-wris}
Q.~Dai, S.~Yang, Curriculum point prompting for weakly-supervised referring image segmentation, in: 2024 IEEE/CVF Conference on Computer Vision and Pattern Recognition (CVPR), 2024, pp. 13711--13722.

\bibitem{CLIP}
A.~Radford, J.~W. Kim, C.~Hallacy, A.~Ramesh, G.~Goh, S.~Agarwal, G.~Sastry, A.~Askell, P.~Mishkin, J.~Clark, et~al., Learning transferable visual models from natural language supervision, in: International conference on machine learning, PMLR, 2021, pp. 8748--8763.

\bibitem{SAM}
A.~Kirillov, E.~Mintun, N.~Ravi, H.~Mao, C.~Rolland, L.~Gustafson, T.~Xiao, S.~Whitehead, A.~C. Berg, W.-Y. Lo, P.~Dollár, R.~Girshick, Segment anything, in: 2023 IEEE/CVF International Conference on Computer Vision (ICCV), 2023, pp. 3992--4003.

\bibitem{Freesolo}
X.~Wang, Z.~Yu, S.~De~Mello, J.~Kautz, A.~Anandkumar, C.~Shen, J.~M. Alvarez, Freesolo: Learning to segment objects without annotations, in: 2022 IEEE/CVF Conference on Computer Vision and Pattern Recognition (CVPR), 2022, pp. 14156--14166.

\bibitem{GLCLIP2023}
S.~Yu, P.~H. Seo, J.~Son, Zero-shot referring image segmentation with global-local context features, in: 2023 IEEE/CVF Conference on Computer Vision and Pattern Recognition (CVPR), 2023, pp. 19456--19465.

\bibitem{TAS2023}
Y.~Suo, L.~Zhu, Y.~Yang, Text augmented spatial aware zero-shot referring image segmentation, in: Findings of the Association for Computational Linguistics: EMNLP 2023, 2023, pp. 1032--1043.

\bibitem{BMS}
W.~Li, C.~Pang, W.~Nie, H.~Tian, A.-A. Liu, Bidirectional mask selection for zero-shot referring image segmentation, IEEE Transactions on Circuits and Systems for Video Technology (2024) 1--1.

\bibitem{blip2}
J.~Li, D.~Li, S.~Savarese, S.~Hoi, Blip-2: Bootstrapping language-image pre-training with frozen image encoders and large language models, in: International conference on machine learning, PMLR, 2023, pp. 19730--19742.

\bibitem{clipasrnn}
S.~Sun, R.~Li, P.~Torr, X.~Gu, S.~Li, Clip as rnn: Segment countless visual concepts without training endeavor, in: 2024 IEEE/CVF Conference on Computer Vision and Pattern Recognition (CVPR), 2024, pp. 13171--13182.

\bibitem{zrec1}
J.~Li, G.~Shakhnarovich, R.~A. Yeh, Adapting clip for phrase localization without further training, arXiv preprint arXiv:2204.03647 (2022).

\bibitem{zrec2}
X.~Liu, S.~Huang, Y.~Kang, H.~Chen, D.~Wang, Vgdiffzero: Text-to-image diffusion models can be zero-shot visual grounders, in: ICASSP 2024-2024 IEEE International Conference on Acoustics, Speech and Signal Processing (ICASSP), IEEE, 2024, pp. 2765--2769.

\bibitem{EFN}
G.~Feng, Z.~Hu, L.~Zhang, H.~Lu, Encoder fusion network with co-attention embedding for referring image segmentation, in: 2021 IEEE/CVF Conference on Computer Vision and Pattern Recognition (CVPR), 2021, pp. 15501--15510.

\bibitem{llava}
H.~Liu, C.~Li, Q.~Wu, Y.~J. Lee, Visual instruction tuning, Advances in neural information processing systems 36 (2024).

\bibitem{gpt4}
D.~Zhu, J.~Chen, X.~Shen, X.~Li, M.~Elhoseiny, Minigpt-4: Enhancing vision-language understanding with advanced large language models, arXiv preprint arXiv:2304.10592 (2023).

\bibitem{lisa}
X.~Lai, Z.~Tian, Y.~Chen, Y.~Li, Y.~Yuan, S.~Liu, J.~Jia, Lisa: Reasoning segmentation via large language model, in: Proceedings of the IEEE/CVF Conference on Computer Vision and Pattern Recognition, 2024, pp. 9579--9589.

\bibitem{gsva}
Z.~Xia, D.~Han, Y.~Han, X.~Pan, S.~Song, G.~Huang, Gsva: Generalized segmentation via multimodal large language models, in: Proceedings of the IEEE/CVF Conference on Computer Vision and Pattern Recognition, 2024, pp. 3858--3869.

\bibitem{PixelLM}
Z.~Ren, Z.~Huang, Y.~Wei, Y.~Zhao, D.~Fu, J.~Feng, X.~Jin, Pixellm: Pixel reasoning with large multimodal model, in: Proceedings of the IEEE/CVF Conference on Computer Vision and Pattern Recognition (CVPR), 2024, pp. 26374--26383.

\bibitem{chen2025sam4MLLM}
Y.-C. Chen, W.-H. Li, C.~Sun, Y.-C.~F. Wang, C.-S. Chen, Sam4mllm: Enhance multi-modal large language model for referring expression segmentation, in: European Conference on Computer Vision, Springer, 2025, pp. 323--340.

\bibitem{zhang2024omg}
T.~Zhang, X.~Li, H.~Fei, H.~Yuan, S.~Wu, S.~Ji, C.~C. Loy, S.~Yan, Omg-llava: Bridging image-level, object-level, pixel-level reasoning and understanding, arXiv preprint arXiv:2406.19389 (2024).

\bibitem{coop}
K.~Zhou, J.~Yang, C.~C. Loy, Z.~Liu, Learning to prompt for vision-language models, International Journal of Computer Vision 130~(9) (2022) 2337--2348.

\bibitem{cocoop}
K.~Zhou, J.~Yang, C.~C. Loy, Z.~Liu, Conditional prompt learning for vision-language models, in: 2022 IEEE/CVF Conference on Computer Vision and Pattern Recognition (CVPR), 2022, pp. 16795--16804.

\bibitem{distillation1}
J.~Ding, N.~Xue, G.~Xia, D.~Dai, Decoupling zero-shot semantic segmentation, in: 2022 IEEE/CVF Conference on Computer Vision and Pattern Recognition (CVPR), 2022, pp. 11573--11582.

\bibitem{ALIGN}
C.~Jia, Y.~Yang, Y.~Xia, Y.-T. Chen, Z.~Parekh, H.~Pham, Q.~Le, Y.-H. Sung, Z.~Li, T.~Duerig, Scaling up visual and vision-language representation learning with noisy text supervision, in: International conference on machine learning, PMLR, 2021, pp. 4904--4916.

\bibitem{stt}
J.~Jang, C.~Kong, D.~Jeon, S.~Kim, N.~Kwak, Unifying vision-language representation space with single-tower transformer, in: Proceedings of the AAAI Conference on Artificial Intelligence, Vol.~37, 2023, pp. 980--988.

\bibitem{Alpha-CLIP}
Z.~Sun, Y.~Fang, T.~Wu, P.~Zhang, Y.~Zang, S.~Kong, Y.~Xiong, D.~Lin, J.~Wang, Alpha-clip: A clip model focusing on wherever you want, in: 2024 IEEE/CVF Conference on Computer Vision and Pattern Recognition (CVPR), 2024, pp. 13019--13029.

\bibitem{spacy}
M.~Honnibal, M.~Johnson, An improved non-monotonic transition system for dependency parsing, in: Proceedings of the 2015 conference on empirical methods in natural language processing, 2015, pp. 1373--1378.

\bibitem{refcoco}
V.~K. Nagaraja, V.~I. Morariu, L.~S. Davis, Modeling context between objects for referring expression understanding, in: Computer Vision--ECCV 2016: 14th European Conference, Amsterdam, The Netherlands, October 11--14, 2016, Proceedings, Part IV 14, Springer, 2016, pp. 792--807.

\bibitem{refcocog1}
S.~Kazemzadeh, V.~Ordonez, M.~Matten, T.~Berg, Referitgame: Referring to objects in photographs of natural scenes, in: Proceedings of the 2014 conference on empirical methods in natural language processing (EMNLP), 2014, pp. 787--798.

\bibitem{refcocog2}
J.~Mao, J.~Huang, A.~Toshev, O.~Camburu, A.~L. Yuille, K.~Murphy, Generation and comprehension of unambiguous object descriptions, in: Proceedings of the IEEE conference on computer vision and pattern recognition, 2016, pp. 11--20.

\bibitem{Grad-CAM}
R.~R. Selvaraju, M.~Cogswell, A.~Das, R.~Vedantam, D.~Parikh, D.~Batra, Grad-cam: visual explanations from deep networks via gradient-based localization, International journal of computer vision 128 (2020) 336--359.

\bibitem{ScoreMap}
C.~Zhou, C.~C. Loy, B.~Dai, Extract free dense labels from clip, in: European Conference on Computer Vision, Springer, 2022, pp. 696--712.

\bibitem{ClipSurgery}
L.~Xu, M.~H. Huang, X.~Shang, Z.~Yuan, Y.~Sun, J.~Liu, Meta compositional referring expression segmentation, in: Proceedings of the IEEE/CVF Conference on Computer Vision and Pattern Recognition, 2023, pp. 19478--19487.

\bibitem{ssc-wris}
F.~Eiras, K.~Oksuz, A.~Bibi, P.~H. Torr, P.~K. Dokania, Segment, select, correct: A framework for weakly-supervised referring segmentation, arXiv preprint arXiv:2310.13479 (2023).

\bibitem{Groundingdino}
S.~Liu, Z.~Zeng, T.~Ren, F.~Li, H.~Zhang, J.~Yang, Q.~Jiang, C.~Li, J.~Yang, H.~Su, et~al., Grounding dino: Marrying dino with grounded pre-training for open-set object detection, in: European Conference on Computer Vision, Springer, 2025, pp. 38--55.

\bibitem{liu2024improved}
H.~Liu, C.~Li, Y.~Li, Y.~J. Lee, Improved baselines with visual instruction tuning, in: Proceedings of the IEEE/CVF Conference on Computer Vision and Pattern Recognition, 2024, pp. 26296--26306.

\bibitem{bai2023qwen}
J.~Bai, S.~Bai, Y.~Chu, Z.~Cui, K.~Dang, X.~Deng, Y.~Fan, W.~Ge, Y.~Han, F.~Huang, et~al., Qwen technical report, arXiv preprint arXiv:2309.16609 (2023).

\end{thebibliography}

\end{document}